\documentclass{article}

\usepackage{arxiv}

\usepackage[utf8]{inputenc} 
\usepackage[T1]{fontenc}    
\usepackage{hyperref}       
\usepackage{url}            
\usepackage{booktabs}       
\usepackage{amsfonts}       
\usepackage{nicefrac}       
\usepackage{microtype}      
\usepackage{lipsum}
\usepackage{graphicx}
\graphicspath{ {./images/} }
\usepackage{subfig}
\usepackage{amsmath}
\usepackage{multirow}
\usepackage{algorithm}
\usepackage{algpseudocode}
\algtext*{EndWhile}
\algtext*{EndIf}

\title{Intention-aware Hierarchical Diffusion Model for\\ Long-term Trajectory Anomaly Detection}

\author{
Chen Wang\\
School of Computing and Information Systems\\
 The University of Melbourne\\
 Melbourne, Australia\\
 \texttt{cww3@student.unimelb.edu.au}
 \And
Sarah Erfani \\
School of Coumputing and Information\\
The University of Melbourne\\
Melbourne, Australia\\
\texttt{sarah.erfani@unimelb.edu.au}
\And
Tansu Alpcan \\
Department of Electrical and Electronic Engineering\\
The University of Melbourne\\
Melbourne, Australia\\
  \texttt{tansualpcan@unimelb.edu.au} 
  \And
 Christopher Leckie \\
  School of Coumputing and Information\\
  The University of Melbourne\\
 Melbourne, Australia\\
  \texttt{caleckie@unimelb.edu.au} \\
}

\begin{document}
\maketitle
\begin{abstract}
Long-term trajectory anomaly detection is a challenging problem due to the diversity and complex spatiotemporal dependencies in trajectory data. Existing trajectory anomaly detection methods fail to simultaneously consider both the high-level intentions of agents as well as the low-level details of the agent's navigation when analysing an agent's trajectories. This limits their ability to capture the full diversity of normal trajectories. 
In this paper, we propose an unsupervised trajectory anomaly detection method named \textbf{I}ntention-aware \textbf{Hi}erarchical \textbf{D}iffusion model (IHiD), which detects anomalies through both high-level intent evaluation and low-level subtrajectory analysis. Our approach leverages Inverse Q Learning as the high-level model to assess whether a selected subgoal aligns with an agent’s intention based on predicted Q-values. Meanwhile, a diffusion model serves as the low-level model to generate subtrajectories conditioned on subgoal information, with anomaly detection based on reconstruction error.
By integrating both models, IHiD effectively utilises subgoal transition knowledge and is designed to capture the diverse distribution of normal trajectories. Our experiments show that the proposed method IHiD achieves up to $30.2\%$ improvement in anomaly detection performance in terms of $F_1$ score over state-of-the-art baselines. 
\end{abstract}


\section{Introduction}
Long-term trajectories refer to spatial-temporal sequences over extended periods and usually involve richer contextual information compared to short-term or immediate paths. 
Long-term trajectories are common in a wide range of real-world activities such as traffic management \cite{saleh2018long} and maritime monitoring \cite{zhang2024long}. With the availability of large-scale data, long-term trajectory anomaly detection has become a significant task in spatial-temporal data analysis.
Traditional machine learning methods suffer from a significant performance drop in long-term trajectories \cite{yao2022trajgat} as they fail to capture diversity and deal with uncertainty in such trajectories. 

\begin{figure}[t]
  \centering
  \includegraphics[width=0.7\linewidth]{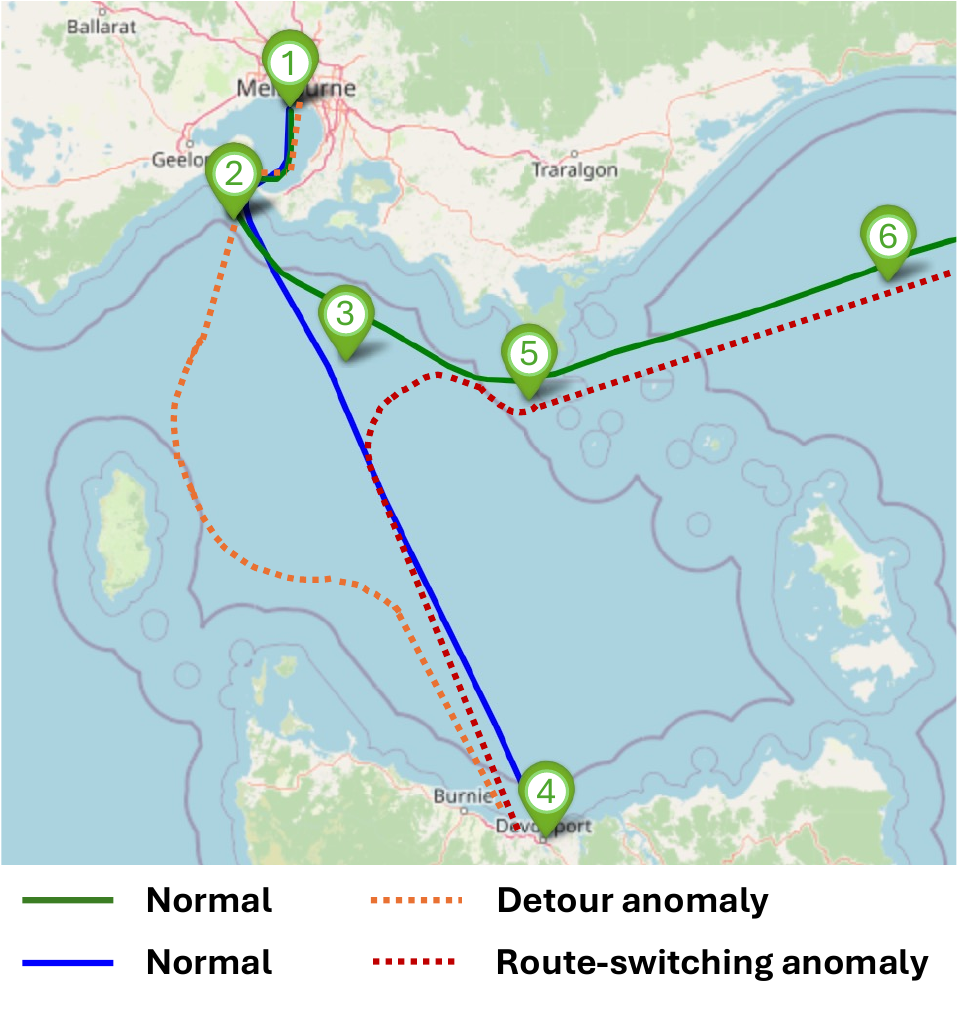}
  \caption{Examples of trajectory anomaly detection from the AIS dataset. Each long-term trajectory is segmented to a series of subtrajectories based on subgoal nodes.}
  \label{fig:intro}
\end{figure}

Long-term trajectories of goal-oriented agents, such as humans, vehicles, vessels, or robots, can often be viewed as a sequence of subgoals, each associated with a corresponding subtrajectory to achieve it \cite{liang2023hierarchical,xu2023bits,seo2024idil}. 
In this framework, an anomalous trajectory indicates either an unusual subgoal selection or a different way to accomplish the chosen subgoal. Consider the examples in Figure \ref{fig:intro}. Normal trajectories, represented in blue and green, can be decomposed into a specific sequence of subtrajectories. The green trajectory follows the subgoals $1\rightarrow2\rightarrow5\rightarrow6$, while the blue trajectory adheres to the subgoals $4\rightarrow3\rightarrow2\rightarrow1$.
By contrast, the red dotted trajectory is considered as a route-switching anomaly because it transitions from one normal route to another - specifically, from subgoal 4 to subgoal 5. The detour anomaly (orange) exhibits a deviation from the normal route to between subgoals 4 and 2. 
\begin{figure*}[t]
  \centering
  \includegraphics[width=\linewidth]{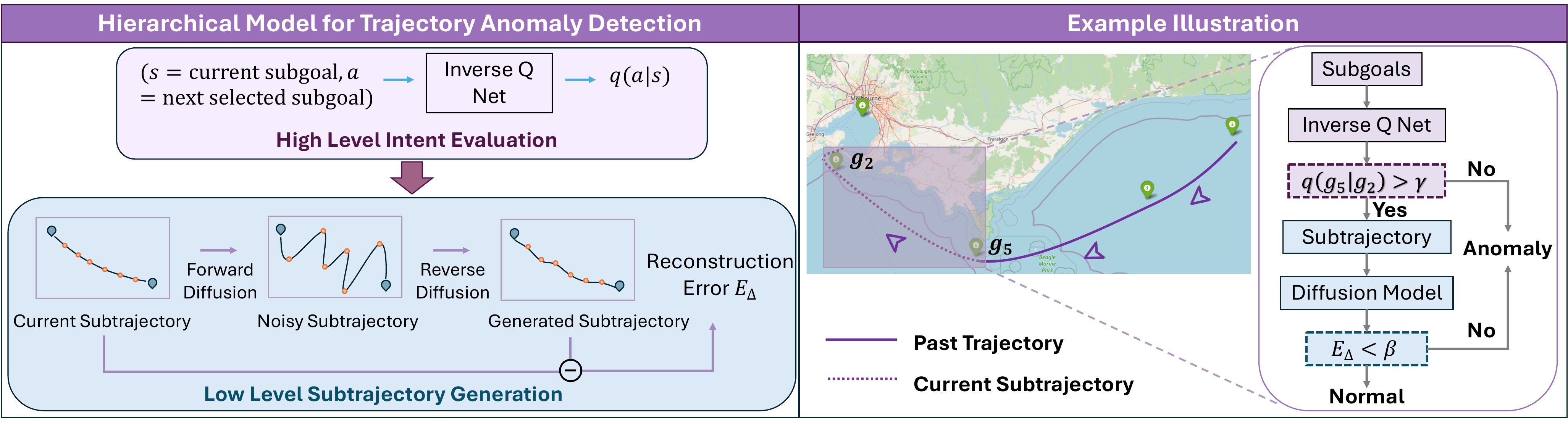}
  \caption{Method overview. The hierarchical model includes two stages: (1) high-level intent evaluation to detect anomalous subgoal selection and (2) low-level subtrajectoy generation to detect any anomalous route deviation. $\gamma$ and $\beta$ are pre-defined thresholds.}
  \label{fig:method}
\end{figure*}

There are two main challenges in long-term trajectory anomaly detection. 
\textbf{(1) Diversity of trajectories}: Long-term trajectories with similar pasts may diverge at certain points, leading to different paths. Furthermore, there are multiple possible destinations for normal trajectories and several valid routes exist even between the same source-destination pairs. 
\textbf{(2) Spatial-temporal consistency}: Capturing long-term dependency in a large spatial-temporal dataset is challenging. Most existing methods struggle with preserving long-term dependencies, and this leads to abrupt transitions or unrealistic behaviour within the trajectory.

Traditional trajectory anomaly detection methods are mainly deterministic \cite{zhang2011ibat,chen2013iboat,liu2013density}, which limits their ability to capture the diversity of multiple possible normal routes. To address this issue, modern approaches have shifted to generative models to capture trajectory variations probabilistically \cite{han2022deeptea,liu2020online}. While many models incorporate contextual information during trajectory generation, such as past trajectories, time and road networks, the same context can still result in multiple plausible paths.
We argue that the intents driving agents serve as a high-level behavioural representation and offer a more effective approach to modelling the multimodal nature of future trajectories.
Moreover, most deep learning-based approaches struggle to capture long-term dependencies \cite{yao2022trajgat}. To mitigate this issue, some imitation learning methods integrate intent or subgoal information as guidance for model learning \cite{seo2024idil}. However, this has not been explored in trajectory anomaly detection.

In this paper, we propose an unsupervised trajectory anomaly detection framework named \textbf{i}ntention-aware \textbf{hi}erarchical \textbf{d}iffusion model (IHiD), designed to detect long-term anomalous trajectories in highly diverse environments. Our approach identifies anomalies based on both high-level intent and fine-grained differences in sub-trajectories. Specifically, we assume that the overarching goal of long-term trajectories is to reach a destination, with each trajectory consisting of a sequence of subgoals.

To handle the diversity of long-term trajectories and enhance spatial-temporal consistency, we design a hierarchical structure to model the goal-oriented behaviour.
Inspired by how people decompose a complex task into smaller and manageable subgoals, the proposed hierarchical approach generates each individual subtrajectory based on the current subgoal.
For example, in maritime navigation, vessels frequently segment their long-distance voyages into stages defined by intermediate waypoints, to ensure safer and more efficient travel. 

Existing hierarchical models \cite{zeng2021hierarchical, kong2024unbiased} for modelling spatial-temporal trajectories do not consider inferring the underlying intentions of trajectories. To address this gap, we draw inspiration from hierarchical imitation learning approaches \cite{seo2024idil}, which decomposes the learning process into high-level intent inference and low-level action generation. However, unlike controlled settings in imitation learning, real-world anomaly detection rarely allows for online interactions or access to environment dynamics \cite{wang2025oil}. At the same time, fully offline hierarchical frameworks often suffer a performance drop as the number of subgoals increases \cite{jain2024go}.
In contrast to existing hierarchical imitation learning methods, our framework is specifically designed for real-world anomaly detection scenarios, where we assume access only to deterministic subgoal dynamics. This maintains a practical balance—enabling intent-aware modelling without requiring full environment knowledge or interactivity; therefore, our approach is feasible and effective in complex, real-life settings.

Our model structure is shown in Figure \ref{fig:method}. Specifically, for the high-level intent inference, we adopt inverse Q learning (IQL) \cite{garg2021iq}, which aims to learn a Q function from one agent's trajectories (e.g., sequences of subgoals) to enable other agents imitate the demonstrated behaviour. The learned Q function shows the intentions and preferences underlying the observed agent's actions. By inferring the Q function from normal agent behaviours, we can assess whether a new agent’s subgoal choice aligns with typical intents.
If a subgoal selection is anomalous, the Q value for this selected subgoal is lower than the others. 
Meanwhile, for the low-level subtrajectory generation, we employ a diffusion model to generate subtrajectories. 
This is motivated by the success of diffusion models in deep generative models \cite{ho2020denoising, rombach2022high}.
They have also become a strong candidate in anomaly detection due to their superior ability in mode coverage and high-quality samples \cite{li2024difftad}.
The reconstruction error between the generated subtrajectory and the real trajectory is used as a second detection.


The main contributions of our work are as follows:
\begin{itemize}
    \item We propose a novel hierarchical framework named IHiD for trajectory anomaly detection. To the best of our knowledge, this is the first work to integrate IQL and diffusion models within a hierarchical framework.
    \item Our model effectively preserves long-term spatial-temporal consistency by considering uncertainty in agents' intentions (subgoals) and incorporating a subgoal-based diffusion model for subtrajectory generation.
    \item Our method can detect anomalies in highly diverse datasets, where normal trajectories can vary significantly even within the same source-destination pair.
    \item Experimental results demonstrate that IHiD achieves an average improvement of approximately $18\%$ and a maximum improvement of $30.2\%$ in $F_1$ score compared to state-of-the-art methods.
\end{itemize}

\section{Related Work}
In this section, we provide an overview of related studies from three perspectives: (1) Hierarchical models for sequential anomaly detection; (2) IRL-based intent inference; and (3) Diffusion-based anomaly detection.
\subsection{Hierarchical Models for Sequential Anomaly Detection}

Hierarchical models have been applied to sequential anomaly detection due to its ability to capture complex dependencies. 
  Temporal Hierarchical One-Class (THOC) network \cite{shen2020timeseries} employs a hierarchical clustering mechanism to fuse information from all intermediate layers of an RNN, rather than relying solely on the final layer.
InterFusion \cite{li2021multivariate} utilises a hierarchical Variational Auto-Encoder to reconstruct high-dimensional input data with two latent variables, representing intermetric and temporal dependency, respectively. While these studies focus on detecting anomalies in multivariate time series, our work centres on trajectories with spatial-temporal information.

To detect trajectory anomalies, HSTGCNN \cite{zeng2021hierarchical} uses a hierarchical spatial-temporal graph convolutional neural network to model human activities in videos with a high-level graph representation to encode interactions among observed people, and a low-level graph representation to encode each person's movement. HUAD (Hierarchical Urban Anomaly Detection) \cite{kong2020huad} generates refined anomaly characteristics with two One-Class Support Vector Machine (OCSVM). 
HS-UATD \cite{kong2024unbiased} is a hierarchical sequence modelling method that partitions the entire spatial region into grids of varying sizes using a density-based quadtree. The hierarchical structure of the quadtree is then leveraged to capture the similarities and differences among trajectories.
However, these methods overlook the critical aspect of inferring the underlying intentions of trajectories, despite the fact that human driving behaviour is typically goal-driven and intentional. Understanding these intentions is essential, as they provide valuable context for distinguishing normal behaviours from anomalies, enabling more accurate and meaningful trajectory analysis.

\subsection{IRL-based Intent Inference}
IRL and Imitation Learning have been widely used to learn preferences from behavioural data and explain human decision-making \cite{pan2020xgail}. For example, in \cite{feng2023ilroute}, a reward function is learned based on personalised constraints, mobility regularities and long-term rewards to determine the best routing strategy for food delivery services. A learned reward function by IRL can also be used for anomaly detection, as it captures an agent's preferences and provides an explanation of its normal behaviour \cite{oh2019sequential}.

Since intents influence expert behaviour on a macro scale, learning intent-driven behaviour is closely related to hierarchical imitation learning \cite{seo2024idil}.
Given that human driving behaviour is usually intentional, several studies have employed hierarchical imitation learning frameworks that decompose the problem into high-level intent inference and low-level action generation \cite{xu2023bits}. 
However, these methods often require known environment dynamics or online interactions with the environment at both high- and low-level sequences \cite{jing2021adversarial}. Additionally, entirely offline frameworks struggle to maintain performance as the diversity of the data increases \cite{jain2024go}. 

Building on these advances, we adopt a hierarchical imitation learning framework that incorporates a high-level model for subgoal evaluation and a low-level goal-conditioned model for action generation. Our framework differs from existing approaches in two key aspects: (1) it is specifically designed for anomaly detection; (2) it only requires known dynamics for subgoal transitions; (3) the low-level model is implemented as a diffusion model to reconstruct diverse action trajectories effectively.
\subsection{Diffusion-based Anomaly Detection}
Diffusion models have an advantage over other generative models in learning complex and multimodal distributions. Recently, many works have successfully applied diffusion models to anomaly detection. Most existing approaches rely on reconstruction errors derived from trained diffusion models to detect anomalies \cite{wyatt2022anoddpm,pintilie2023time,liu2023unsupervised,li2024difftad}. They typically involve a forward-backward diffusion process, where data is first corrupted with noise and then reconstructed. Diffusion models trained on normal data learns to reconstruct samples based on the underlying distribution of the normal data. Therefore, large differences between the original and reconstructed data act as indicators of anomalous behaviour.
Prediction-based methods train a diffusion model conditioned on past frames to estimate future frames. 
MoCoDAD \cite{flaborea2023multimodal} detects anomalies based on the observation that the true future motions often lie in the marginal regions of the generated sample distribution when the model is conditioned on anomalous past motions. In order to reduce the expensive inference time, another work DTE estimates the posterior distribution of diffusion time (noise variance) for a given noisy input, showing that the diffusion time is linked to the distance between the noisy input and its corresponding denoised reconstruction \cite{livernoche2023diffusion}. An imputation-based anomaly detection method DiffAD \cite{xiao2023imputation} uses a diffusion model conditioned on observed points (normal data) to estimate masked points, and identifies anomalies by comparing estimation errors. Another work DeCoRTAD \cite{wang2024decortad} integrates a diffusion model with two encoders to learn a representation space for trajectory anomaly detection.

Our low-level diffusion model operates within a reconstruction-based detection framework. By focusing on reconstructing sub-trajectories, it partially mitigates the high computational cost associated with diffusion models.

\section{Preliminaries}
In this section, we give a brief introduction to Markov Decision Processes, Inverse Q Learning and diffusion models. 
\subsection{Markov Decision Processes}
Markov Decision Processes (MDPs) provide a mathematical foundation for reinforcement learning (RL). An MDP is defined by the tuple $(S,A,\mathcal{P},\mathcal{R},\gamma,b_0)$, where $S$ is a state space, $A$ is an action space, $\mathcal{P}_{ss'}^a = P(S_{t+1}=s'|S_t=s,A_t=a)$ is the transition probability, $\mathcal{R}_s^a=\mathbb{E}[R_{t+1}|S_t=s,A_t=a]$ represents the expected reward, $\gamma \in [0,1)$ is the discount factor and $b_0(s)$ specifies the probability of starting in state $s$. 
At each state $s$, the agent selects an action $a$, and the environment transitions to a new state $s'$ based on the transition probability $\mathcal{P}$. Simultaneously, the agent receives a corresponding reward $\mathcal{R}_s^a$. The objective of an MDP is to determine an optimal policy $\pi^*$ that maximizes the expected discounted sum of future rewards. A policy $\pi$ defines a probability distribution over actions for a given state, expressed as $\pi (a|s) = P[A_t = a|S_t = s]$. Each policy is associated with a state-value function: $v_\pi (s)=\mathbb{E}_{\pi}[\mathcal{R}_{t+1}+\gamma v_\pi(S_{t+1})|S_t=s]$ and a Q-value function: $Q_\pi (s,a) = \mathbb{E}_{\pi}[\mathcal{R}_{t+1}+\gamma Q_\pi(S_{t+1},A_{t+1})|S_t=s, A_t=a]$. The optimal policy $\pi^*$ can be derived by maximizing the action-value function, where $Q_* (s,a)=\max_{\pi} Q_{\pi}(s,a)$. 
\subsection{Inverse Reinforcement Learning (IRL) and Inverse Q Learning (IQL)}
IRL has emerged as an important research area in recent years, gaining significant attention for its ability to estimate the reward function from experts’ demonstrations \cite{ziebart2008maximum}. However, IRL faces several challenges, particularly in adversarial-based methods \cite{fu2017learning}, where the learning process can be unstable due to the min-max formulation over the reward and policy \cite{garg2021iq}.
Given expert demonstrations sampled from expert policy $\pi_E$, maximum entropy IRL seeks to recover the reward function $r$ from a set of functions $\mathcal{R}$. Simultaneously, it searches for the best policy $\pi$ for the reward function in an inner loop.
The optimisation objective of this process is formulated as:
\begin{equation}
    \max_{r\in\mathcal{R}} \min_{\pi\in\Pi}L(\pi,r)=\mathbf{E}_{\rho_E}[r(s,a)]-\mathbf{E}_{\rho_\pi}[r(s,a)]-H(\pi)
\end{equation}
where $H(\pi)=\mathbf{E}_{\rho_\pi}[-\log\pi(a|s)]$ is the discounted causal entropy of the policy $\pi$.

Unlike conventional IRL methods, IQL learns Q-functions to recover expert behaviour, instead of learning a reward function or policy \cite{garg2021iq}. This framework simplifies the unstable min-max problem involving policies and reward functions into a straightforward minimisation problem focused on a single Q-function:
\begin{equation}
    \max_Q \mathbf{E_{\rho_E}}[\mathcal{T}[Q](s,a)]-(1-\gamma)\mathbf{E}_{s_0}[V^Q(s)]-\psi(\mathcal{T}[Q](s,a))
    \label{equ:obj_iql}
\end{equation}
where $\mathcal{T}[Q](s,a)=Q(s,a)-\gamma\mathbf{E}_{s'\sim P(s'|s,a)}[V^Q(s)]$, $V^Q(s)$ is a softmax of the Q function $V^Q(s)=\log(\sum_a \exp (Q(s,a)))$, and $\psi(\cdot)$ is a convex regulariser.
In MDPs, Q-functions represent the expected cumulative reward an agent can achieve by starting from a given state and taking a specific action. They serve as a crucial link between the reward function and the optimal policy in a given environment.
\subsection{Diffusion Models}
Denoising Diffusion Probabilistic Models (DDPMs), or diffusion models, are state-of-the-art generative models excelling in sample quality and mode coverage \cite{ho2020denoising,rombach2022high}. They consist of two key processes: a forward diffusion process that gradually adds Gaussian noise to the data and a reverse process that learns to remove it step-by-step, enabling high-quality sample generation.

The forward diffusion process is a Markov chain of $T$ steps. For input data $x^0$ sampled from data distribution $q(x)$, noise is added at each step $t$ with variance $\beta^t$,
\begin{equation}
    q(x^t|x^{t-1}) = \mathcal{N}(x^t|x^{t-1}\sqrt{1-\beta^t}, \beta^t\mathbf{I}).
\end{equation}
Using $\alpha^t=1-\beta^t$ and $\bar{\alpha}^t=\prod_{s=0}^t\alpha^s$, $x^t$ can be directly sampled as:
\begin{equation}
    x^t \sim q(x^t|x^{0}) = \mathcal{N}(x^t|\sqrt{\bar{\alpha}^t}x^0, (1-\bar{\alpha}^t)\mathbf{I}).
\end{equation}

The reverse process seeks to approximate the reverse distribution $p_{\theta}(x^{t-1}|x^t)$ via a neural network parameterise by $\theta$:
\begin{equation}
    p_{\theta}(x^{t-1}|x^t) = \mathcal{N}(x^{t-1}|\mu_{\theta}(x^t,t),\Tilde{\beta}^t\mathbf{I}),
\end{equation}
where $\Tilde{\beta}^t=\frac{1-\bar{\alpha}^{t-1}}{1-\bar{\alpha}^{t}}\beta^t$. The mean $\mu_{\theta}(x^t,t)$ is parameterized as \cite{ho2020denoising}:
\begin{equation}
    \mu_{\theta}(x^t,t)=\frac{1}{\sqrt{\alpha^t}}(x^t-\frac{\beta^t}{\sqrt{1-\bar{\alpha}^t}}\epsilon_\theta(x^t,t)),
\end{equation}
where $\epsilon_\theta(x^t,t)$ is learned by a neural network. Samples are generated iteratively from $x^T$ to $x^0$ using:
\begin{equation}
    x^{t-1}=\frac{1}{\sqrt{\alpha^t}}(x^t-\frac{\beta^t}{\sqrt{1-\bar{\alpha}^t}}\epsilon_\theta(x^t,t))+\sqrt{\Tilde{\beta}^t}z, \quad z\sim\mathcal{N}(0,\mathbf{I})
\end{equation}

The model's objective is to minimize the negative log-likelihood of the training data. The simplified denoising term in the Evidence Lower Bound (ELBO) is expressed as:
\begin{equation}
    \mathcal{L}_{simple}=\mathbb{E}_{t\sim[1,T],x^0\sim q(x^0),\epsilon\sim\mathcal{N}(0,\mathbf{I})}\|\epsilon-\epsilon_\theta(x^t,t) \|^2.
\end{equation}

Diffusion models often employ UNet-based architectures, originally developed for image processing tasks \cite{ronneberger2015u}. However, these architectures are not inherently suitable for sequential data. To address this limitation, specialized variations have been proposed. For example, Motion Latent-based Diffusion model (MLD) \cite{chen2023executing} integrated transformers into the UNet framework, enhancing its ability to handle sequential data sets effectively.

\section{IHiD: Intention-aware Hierarchical Diffusion}
In this section, we first introduce our hierarchical framework. Then, we explain each component of our model in detail. 
Finally, we show how to detect anomalies following a hierarchical procedure.
\begin{figure*}[t]
  \centering
  \includegraphics[width=0.75\linewidth]{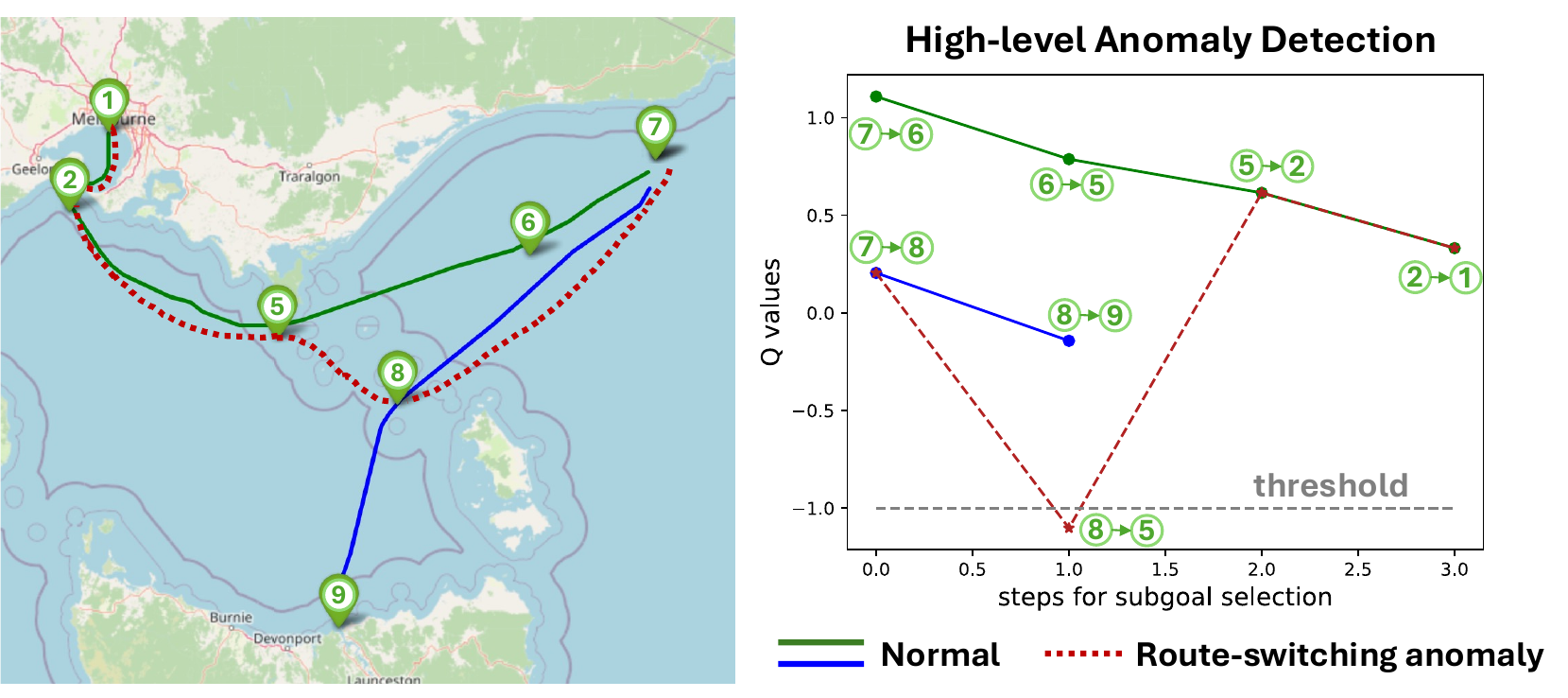}
  \caption{An example of using high-level intent evaluation for anomaly detection.}
  \label{fig:high_level_detect}
\end{figure*}
\subsection{Problem Statement}
We aim to detect anomalies within long-term trajectories. Let $\tau$ denote a trajectory. We first give definitions of related concepts.

\textbf{Trajectory}:
A trajectory is a sequence of time points, $\tau = \{x_1, x_2, \dots, x_T\}$, where each $x_t \in \mathbb{R}^d$ represents spatiotemporal information at time $t$. In this work, we consider $x_t$ to be a two-dimensional location feature with latitude and longitude $x_t=(lat_t, lon_t)$.

\textbf{Long-term Trajectory}:
A long-term trajectory $\tau$ can be decomposed into a subgoal trajectory $\phi=\{g_1, g_2, \dots, g_k\}$ at a high level and $k$ low-level subtrajectories $\{\hat{\tau}_1, \hat{\tau}_2, \dots, \hat{\tau}_k\}$ corresponding to each subgoal $g_i \in \phi$, where $\tau=\{\hat{\tau}_1, \hat{\tau}_2, \dots, \hat{\tau}_k\}$.

In this paper, we consider a diverse trajectory database $\mathcal{T}$ containing $N$ trajectories with $M$ destinations. To reach a destination $d_m$, multiple different subgoal trajectories $\phi$ exist. To reach one subgoal $g$, there are also multiple possible subtrajectories $\hat{\tau}$.
Our objective is to develop a hierarchical framework for long-term trajectory anomaly detection. We assume that the subgoal transition dynamics is known. During the testing stage, we assume that we do not know the final destination $d_m$ for each testing trajectory.

\subsection{Hierarchical Framework}
As shown in Figure \ref{fig:method}, the overall hierarchical framework consists of two main components. 
\textbf{(1) High-Level Model (Inverse Q Net)}: This model evaluates the selected subgoal, defined as the endpoint of a subtrajectory. It uses the current state (starting point of the subtrajectory) and the chosen action (next subgoal) as inputs to predict a Q-value. This Q-value assesses the expected future rewards of selecting the subgoal, thereby determining whether the chosen subgoal aligns with the agent's intention. \textbf{(2) Low-Level Diffusion Model}: This component generates subtrajectories using the contextual information provided by subgoals. By sampling generated subtrajectories from the same data distribution as normal trajectories, reconstruction errors between the current and generated subtrajectories serve as a measure for detecting anomalies. Specifically, larger reconstruction errors indicate a higher likelihood of anomalous behavior.

The training processes of two models are done separately. The high-level model is trained to predict the next subgoal using the subgoal trajectories, and the low-level model is trained to reconstruct the low-level subtrajectories conditioned on a specific subgoal.
\subsection{Graph Construction}
To facilitate subgoal identification, we constructed a graph for each dataset using the training data. Nodes in the graph were defined based on the following rules: (1) all destinations are designated as nodes; (2) turning points with a frequency exceeding a predefined threshold in the training trajectories are included as nodes; and (3) the distance between any two nodes must exceed a predefined threshold. Two nodes are connected if a corresponding route exists in the training trajectories. These nodes in the graph represent the subgoals of agents.
To reduce computational costs, not all turning points are included in the graph; instead, we prioritise those with higher frequencies.

\begin{algorithm}[t]
\caption{Training Process for High-level Inverse Q Net}\label{alg:train}
 \begin{algorithmic}[1]
\renewcommand{\algorithmicrequire}{\textbf{Input:}}
\renewcommand{\algorithmicensure}{\textbf{Output:}}
\Require High-level subgoal trajectories G
\Repeat
\State $\phi \sim G$ 
\State $(g_i, g_{i+1}) \sim\phi$
\State state $s=g_i$
\State action $a=g_{i+1}$
\State next state $s'=g_{i+1}$
\State Train Q function based on Equation \ref{equ:obj_iql}

\Until{converged}
\end{algorithmic}
\end{algorithm}
\subsection{High-level Intent Evaluation}


As intents often influence agents' behaviours at a macro level, various studies incorporate latent states, such as subtasks, subgoals, strategies, or mental states, to learn intent-driven behaviour \cite{seo2024idil,jain2024go,jing2021adversarial,kipf2019compile}. In this work, we adopt the concept of subgoals from hierarchical imitation learning to capture the diverse trajectory patterns.

We consider subgoal trajectories within an MDP environment, defined by the tuple $(S,A,\mathcal{P},\mathcal{R},\gamma,b_0)$ where the state and the action spaces share the same subgoal space.
A high-level decision-making sequence, composed of state-action pairs, is defined as $\{(g_1, g_2), (g_2, g_3), \dots, (g_{k-1},g_k)\}$, meaning that at state $g_1$, the agent selects subgoal $g_2$ as the next target, and so on. We assume a deterministic environment with a known transition probability $\mathcal{P}$, such that $\mathcal{P}(S_{t+1}=g_j|S_t=g_i,A_t=g_j)=1$. The reward function $\mathcal{R}$ is considered unknown. 

We utilise the Inverse soft-Q learning algorithm \cite{garg2021iq} as the high-level model to learn a Q function from subgoal trajectories. The training process is shown in Algorithm \ref{alg:train}.
The learned Q function can evaluate whether a new subgoal in a testing trajectory follows a normal pattern. Specifically, if the Q value for a new subgoal selection is low, it indicates that the subgoal is inconsistent with the incentives of normal agents and has a higher likelihood of being an anomaly. 

Figure \ref{fig:high_level_detect} illustrates the use of Q values for high-level anomaly detection. Two normal trajectories (green $\phi_g$ and blue $\phi_b$) share the same source but have different destinations. The subgoal trajectories for these normal trajectories are $\phi_g = \{g_1=7, g_2=6, g_3=5, g_4=2, g_5=1\}$ and $\phi_b = \{g_1=7, g_2=8, g_3=9\}$, respectively. The route-switching anomaly follows the subgoal trajectory $\phi_{rs}= \{g_1=7, g_2=8, g_3=5, g_4=2, g_5=1\}$. The Q value at the second step for this anomaly falls below the pre-defined threshold, as $Q(S_2=8, A_2=5)<-1$, leading to its detection as an anomaly.

\subsection{Low-level Subtrajectory Generation}
\begin{figure}[t]
  \centering
  \includegraphics[width=0.7\linewidth]{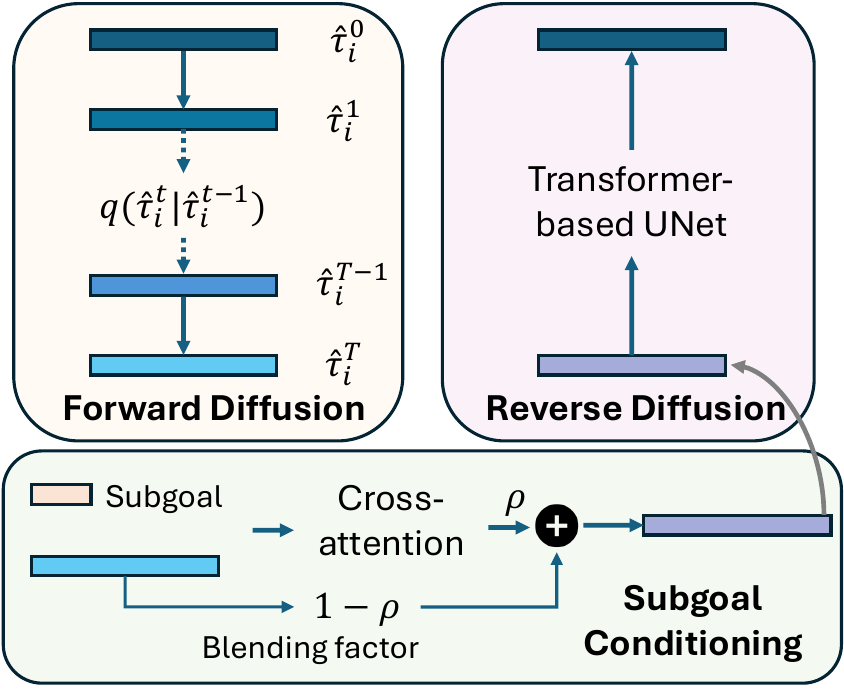}
  \caption{Overview of the diffusion model. During training, the Forward Diffusion adds noise to the subtrajectory gradually with the increasing diffusion step. Then the Reverse Diffusion learns to recover the original subtrajectory by conditioning on the subgoal information. }
  \label{fig:diff}
\end{figure}
\begin{algorithm}[t]
\caption{Training Process for Low-level Diffusion Model}\label{alg:train2}
 \begin{algorithmic}[1]
\renewcommand{\algorithmicrequire}{\textbf{Input:}}
\renewcommand{\algorithmicensure}{\textbf{Output:}}
\Require Normal trajectories $\mathcal{T}$, subgoal trajectories G
\Repeat
\State $\hat{\tau}_i\sim\mathcal{T}$, $(g_i, g_{i+1})\sim G$
\State $\hat{\tau}_i = \hat{\tau}_i^0$
\State $t \sim$ Uniform$(\{1,2,\dots,T-1,T\})$
\State Get $\hat{\tau}_i^t$ based on Equation \ref{equ:addnoise} (after $t$ steps of noise)
\State Compute $c_i$ based on Equation \ref{equ:ci}
\State $\epsilon \sim \mathcal{N}(0,1)$
\State Take gradient descent step on: 
 $\nabla_{\theta}\|\epsilon - \epsilon_\theta(c_i,t)\|^2$
\Until{converged}
\end{algorithmic}
\end{algorithm}
The training process is shown in Algorithm \ref{alg:train2}.
The original UNet architecture of diffusion models is used on image data \cite{ho2020denoising}. To adapt the architecture to fit the sequential nature of trajectories, we did a few modifications based on \cite{chen2023executing}. This model uses transformer layers in the network with long skip connections. Unlike \cite{chen2023executing}, we use a cross-attention layer with blending to incorporate context, where the query comes from subgoals (context) and the keys/values are from subtrajectories. This approach is preferred because the dimensionality of a subgoal is much smaller than that of a subtrajectory, and directly concatenating them may lead to ineffective context representation. 

In the forward diffusion process, the diffusion model adds Gaussian noise to a subtrajectory $\hat{\tau}_i^0$ sampled from the original distribution $q(\hat{\tau}_i)$. The Gaussian noise is defined by a variance schedule that increases linearly from $\beta^1=10^{-4}$ to $\beta^T=0.02$.
Then the noised sample $\hat{\tau}_i^t$ can be formulated as:
\begin{equation}
    \hat{\tau}_i^t \sim q(\hat{\tau}_i^t|\hat{\tau}_i^{0}) = \mathcal{N}(\hat{\tau}_i^t|\sqrt{\bar{\alpha}^t}\hat{\tau}_i^0, (1-\bar{\alpha}^t)\mathbf{I}),
    \label{equ:addnoise}
\end{equation}
where $t \in (0,T]$ is the timestep in the Markov chain, $\alpha^t=1-\beta^t$ and $\bar{\alpha}^t=\prod_{s=0}^t\alpha^s$. We set $T=800$.

With the proposed conditioning method, the input to the reverse diffusion process becomes
\begin{equation}
    c_i = \rho CrossAttention(g_i, g_{i+1}, \hat{\tau}_i^t) + (1-\rho) \hat{\tau}_i^t,
    \label{equ:ci}
\end{equation}
where $\rho$ is a blending factor that controls the relative contribution of the subgoal-guided output versus the original noisy subtrajectory. We set $\rho$ as a constant.
During the training process, the diffusion model aims to learn the reverse distribution $p_\theta(\tau_i^{t-1}|\tau_i^t, g_i, g_{i+1})$ by jointly training the transformer-based UNet and the cross-attention layer. The training objective is as follows:
\begin{equation}
    \mathcal{L}(\theta)=\mathbb{E}_{t,\epsilon,c_i}\|\epsilon - \epsilon_\theta(c_i,t)\|^2,
\end{equation}
where $\epsilon \sim \mathcal{N}(0,1)$.

At the inference stage, the trained diffusion model reconstructs subtrajectory $\hat{\tau_i}'$ based on noisy subtrajectory $\hat{\tau_i}^T$ and subgoals $g_i, g_{i+1}$
The reconstruction error between an original subtrajectory $\hat{\tau_i}$ and a denoised subtrajectory $\hat{\tau_i}'$ is computed as:
\begin{equation}
    E_{\Delta} = \frac{1}{N} \|\hat{\tau_i}-\hat{\tau_i}'\|^2 
    \label{equ:recons error}
\end{equation}
where $N$ is the length of $\hat{\tau_i}$.

\subsection{Anomaly Detection}
\begin{algorithm}[t]
\caption{Detection Process}\label{alg:detect}
 \begin{algorithmic}[1]
\renewcommand{\algorithmicrequire}{\textbf{Input:}}
\renewcommand{\algorithmicensure}{\textbf{Output:}}
\Require test subtrajectory $\hat{\tau}_i$($\hat{\tau}_i^0$), subgoals $g_i, g_{i+1}$, trained Q net, trained diffusion model $\epsilon_\theta$, noise level T, threshold $\gamma$ and $\beta$

\State Compute $Q(s=g_i,a=g_{i+1})$
\If{$Q(s=g_i,a=g_{i+1})>\gamma$}
\State $\hat{\tau}_i^{T}\sim \mathcal{N}(\sqrt{\bar{\alpha}^t}\hat{\tau}_i^0, (1-\bar{\alpha}^t)\mathbf{I})$
\State $\hat{\tau}_i^{'}{^{T}} = \hat{\tau}_i^{T}$
\For{$t=T,\dots,1$}
\State $z\sim \mathcal{N}(\mathbf{0},\mathbf{I})$
\State $\hat{\tau}_i^{'}{^{t-1}}=\frac{1}{\sqrt{\alpha^t}}(\hat{\tau}_i^{'}{^{t}}-\frac{\beta^t}{\sqrt{1-\bar{\alpha}^t}}\epsilon_\theta(\hat{\tau}_i^{'}{^{t}},t, c_i))+\sqrt{\Tilde{\beta}^t}z$
\EndFor
\State Get denoised subtrajectory: $\hat{\tau_i}'=\hat{\tau_i}^{'}{^{0}}$ 
\State Compute $E_\Delta$ based on Equation \ref{equ:recons error}
\If{$E_\Delta<\beta$}
\State result: Normal
\Else
\State result: Anomaly
\EndIf
\Else
\State result: Anomaly
\EndIf
\end{algorithmic}
\end{algorithm}
The detection process is shown in Algorithm \ref{alg:detect}.
Given an ongoing trajectory, we can continuously assess whether the current subtrajectory exhibits any anomalous patterns. Our approach employs a two-stage hierarchical evaluation process, leveraging both a high-level inverse Q network and a low-level diffusion-based generative model.

In the first stage, we evaluate the current subgoal selection using the inverse Q network. Specifically, we compute the Q value corresponding to the chosen subgoal and compare it against a predefined anomaly threshold $\gamma$. If the Q value exceeds $\gamma$, the subgoal is deemed consistent with learned behavior, and we proceed to a fine-grained analysis of the subtrajectory. Conversely, if the Q value falls below $\gamma$, the subtrajectory is immediately identified as anomalous, as it suggests a deviation from the expected normal intents.

In the second stage, assuming the subgoal selection passes the first-stage evaluation, we further assess the subtrajectory using a trained diffusion model. Given the subgoal information, the model reconstructs the expected subtrajectory and computes the reconstruction error, denoted as $E_\Delta$. If $E_\Delta>\beta$, where $\beta$ is a predefined threshold, the subtrajectory is considered as an anomaly; otherwise, it is normal.

\subsection{Algorithm Overview}
We proposed a hierarchical framework for modelling trajectories from a behavioural perspective. 
The overarching goal of a trajectory is to reach a destination, a process that can be decomposed into a sequence of subgoals (e.g., visiting intermediate waypoints) and specific paths performed to reach each subgoal.
Several key factors were considered in designing this framework. At the high level, subgoal transitions can be characterized by deterministic transition probabilities, but at the low level, we assume we do not have access to explicit knowledge of the environment dynamics and the precise actions taken. 
To maximize the use of available information, we utilise online IQL for high-level detection since it can leverage the transition probability in the training process and achieve better results in generalization and adaptability to unseen states. 
For modeling low-level subtrajectories, we utilise a diffusion model, which effectively captures trajectory diversity. By doing this, we keep the low-level model offline, eliminating reliance on environment dynamics or transition probabilities, which is a more realistic setting.

\section{Experiments}
In our experiments, we focus on two research questions:  (1) How does our method compare with the baselines in terms of different types of anomalies? (2) How important is each component in the hierarchical structure?
\subsection{Experimental Setup}
The experimental setup includes datasets, anomaly generation, baselines and evaluation metrics.
\subsubsection{Datasets}
We apply our method and other baselines to two real-life datasets. 

\textbf{1) Chengdu Dataset} \cite{didi} collected real-life taxi trajectories recorded within the Second Ring Road of Chengdu, China in November 2016. The GPS data points are sampled at approximately 2 to 4-second intervals, with each record containing five attributes: driver ID, order ID, timestamp, longitude, and latitude. During preprocessing, trajectories are extracted based on order ID to endure continuity. Each trajectory is a two-dimensional sequence, where each point is represented by its latitude and longitude coordinates. To focus on longer trips, we selected trajectories where the distance between the source and destination exceeds 6 kilometers. We then applied MeanShift clustering algorithm to the trajectory destinations and got four high-traffic hotspot clusters for training and testing.

\textbf{2) AIS Dataset} \cite{AIS} is a real-world dataset of sea vessel traffic data from various regions in Australia. Each recorded point represents a vessel position report, containing attributes such as craft ID, longitude, latitude, course, speed, vessel type, vessel sub-type, vessel length, beam, draught, and timestamp. During preprocessing, we selected the trajectories of cargo, tanker, and tug vessels operating in the Bass Strait from July 2020 to February 2022. We segmented continuous AIS records based on destination ports, such as Melbourne, Geelong, Devonport, Burnie, and Beauty Point, as well as predefined boundary constraints. Each trajectory point consists of latitude and longitude.
\subsubsection{Anomaly Generation}
The datasets do not have labeled anomalies. Therefore, following previous studies on trajectory anomaly detection \cite{han2022deeptea,liu2020online,wang2024decortad}, we use two types of generated anomalies to evaluate the detection performance.

\textbf{1) Big detour anomalies} take inefficient or unintended paths compared to normal agents. They are generated by adding deviations to normal trajectories. We define two parameters for the generation process: deviation $d$ and proportion $\omega$. First, a normal trajectory is randomly selected. Then, a segment constituting $\omega$ of the trajectory is replaced with a detour. The detour is characterised by the deviation $d$, which represents the additional distance travelled compared to the original segment. For the Chengdu dataset, we set $d=0.04$ and $\omega=0.6$. For the AIS dataset, we set $d=1.0$ and $\omega=0.6$.

\textbf{2) Small detour anomalies} are generated using a similar procedure to detour anomalies. The key difference is that the deviations of small detour anomalies occur between two subgoals, occupying only a small fraction of the trajectory. In contrast, detour anomalies are not restricted in this way and are more likely to skip subgoal nodes. We use $\omega^*$ to denote the proportion of a subtrajectory between two subgoal nodes. A subtrajectory is randomly selected from a normal trajectory, and $\omega^*$ of it is replaced with a deviation. For the Chengdu dataset, we set $d=0.04$ and $\omega^*=0.6$; for the AIS dataset, we set $d=1.0$ and $\omega^*=0.6$.

\textbf{3) Route-switching anomalies} switch from one normal route to another. They are generated by selecting segments from two different normal trajectories and constructing a short bridge to connect them. Specifically, the first half of one trajectory and the second half of another are combined to form a route-switching anomaly. To ensure that the two normal trajectories are different routes, the spilt points must be at least $\sigma$ kilometres apart. For the Chengdu dataset, we set $\sigma=0.03$. For the AIS dataset, we set $\sigma=1.0$.
\begin{table}[t]
\begin{center}
	\caption{Hyperparameters}
	\label{tab:hyper}
	\begin{tabular}{l|c|c}\toprule
		Hyperparameter & Chengdu & \hspace*{1.7mm} AIS \hspace*{1.7mm} \\ \midrule
            Attention heads  & 4 & 4  \\
            Dropout &0.2 &0.2 \\
            Transformer layers & 7 & 7 \\
            Latent dimension & 256 &128   \\
            t range & 200& 600  \\
            Blending factor $\rho$& 0.4&0.8  \\
            Context attention dim & 64& 64  \\
            Subgoal encoded dim &30 & 30 \\
            Time num channels &4 &4 \\
            Threshold $\gamma$ & -1.2 & -1 \\
            Threshold $E_{\Delta}$ &0.18 & 0.13\\
		 \bottomrule
	\end{tabular}
\end{center}
\end{table}
\subsubsection{Baselines}
\begin{table*}[t]
\begin{center}
\setlength\tabcolsep{2pt}
	\caption{Comparison of our method IHiD with baselines. The columns P, R and $F_1$ represent the precision, recall and $F_1$ score (as $\%$) respectively. For these three metrics, a higher value indicates a better performance.}
	\label{tab:main results}
	\begin{tabular}{l|ccc|ccc|ccc|ccc|ccc|ccc}\toprule
		\textit{Dataset} & \multicolumn{9}{c|}{\textit{Chengdu}} & \multicolumn{9}{c}{\textit{AIS}}  \\ 
  
    \textit{Anomaly} & \multicolumn{3}{c|}{\textit{Big Dtr}} &\multicolumn{3}{c|}{\textit{Small Dtr}}&\multicolumn{3}{c|}{\textit{Switching}}& \multicolumn{3}{c|}{\textit{Big Dtr}} &\multicolumn{3}{c|}{\textit{Small Dtr}}&\multicolumn{3}{c}{\textit{Switching}}\\
    \textit{Metric}  &P& R &$F_1$ &P& R &$F_1$&P& R &$F_1$&P&R&$F_1$ &P&R&$F_1$&P&R&$F_1$ \\ \midrule
    OCSVM &37.5&24.4&29.6&40.4&27.4&32.6&7.75&4.80&5.92&30.1&97.8 & 46.0 & 34.1&27.2&30.2&26.7&\textbf{81.8}&40.2\\
    LOF & 34.8&50.8&41.3&26.3&33.2&29.4&24.8&32.0&27.9&37.7& 69.4& 48.9&42.2&65.0&51.2& 32.8 & 56.6& 41.5\\
    IF & 34.9&95.6&51.1&28.9&80.1&42.5&9.69&26.4&14.1&24.0 & 68.0& 35.5 &15.7&48.5&23.7& 16.4 & 48.8& 24.5\\
    LSTM-AE  & 64.3 & \textbf{100.0} &78.2 &64.5& \textbf{98.0}& 77.8& 55.6 & 65.9& 60.3& 54.9&98.7&70.6& 54.8& \textbf{100.0} &70.8 &44.3&62.8&51.9 \\
    GM-VASE &\textbf{95.9} &68.7 & 80.1 &60.8&4.40 &8.20&80.2&11.9&20.8& 39.5 &50.8& 44.3&5.93&4.71&5.25& 18.8&17.7&18.3\\
    OIL-AD & 57.3&82.6&67.7&58.1 & 88.4&70.1 & 53.9& 70.8&61.2 & 66.5& 78.2 & 71.9 &61.8 &73.8&67.2&63.3 & 69.4 & 66.1\\
    DeCoRTAD & 61.0 & 97.9 & 75.2 &58.9& 89.8 &71.1 & 27.0 & 23.0 & 24.9 & 62.7 & \textbf{99.8}& 77.0 & 61.5&98.1&75.5&54.6 & 78.9& 64.5\\
    \hline 
    Ours &92.0 & 99.8& \textbf{95.8} &\textbf{91.1}&96.6&\textbf{93.7}&\textbf{90.9} & \textbf{91.9} & \textbf{91.4}& 
    
    \textbf{89.4} & 93.0& \textbf{91.1} &\textbf{88.5} &89.8&\textbf{89.1} & \textbf{87.1}& 79.6 & \textbf{83.1}  \\

		  \bottomrule
	\end{tabular}
\end{center}
\end{table*}

\begin{table*}[ht]
\begin{center}
    \setlength\tabcolsep{2pt}
	\caption{Ablation results ($F_1$ score).}
	\label{tab:ablation}
	\begin{tabular}{cc|ccc|ccc|c}\toprule
		\multirow{2}{*}{High-level IQL} & \multirow{2}{*}{Low-level Diff.} & \multicolumn{3}{c|}{Chengdu} & \multicolumn{3}{c|}{AIS} &\multirow{2}{*}{\hspace*{1.7mm} Avg \hspace*{1.7mm}}\\ 
        & & \textit{Big Dtr}& \textit{Small Dtr}& \textit{Switch}&\textit{Big Dtr}& \textit{Small Dtr}&\textit{Switch} \\ \midrule
        \checkmark & & 96.6 & nan/0 & 97.7& 95.0&nan/0&89.3&63.1\\
         & \checkmark &84.2&96.0&55.9 &66.2&90.8&40.9&72.3\\
        \checkmark & \checkmark &95.8&93.7& 91.4& 91.1&89.1&83.1&\textbf{90.7}\\
		 \bottomrule
	\end{tabular}
\end{center}
\end{table*}
We compare our model with an extensive variety of baselines, including classic anomaly detection methods: One-Class Support Vector Machine (OCSVM), Isolation Forest (IF), Local Outlier Factor (LOF) and deep learning models: Long Short Term Memory Autoencoder (LSTM-AE) \cite{malhotra2016lstm}, Gaussian Mixture Variational Sequence AutoEncoder (GM-VSAE) \cite{liu2020online}, DiffTAD \cite{li2024difftad}, OIL-AD \cite{wang2025oil}, DeCoRTAD \cite{wang2024decortad}. DiffTAD and DeCoRTAD are diffusion-based anomaly detection methods, using reconstruction errors and representation learning respectively. OIL-AD is a reinforcement learning-based approach.
\subsubsection{Evaluation Metrics}
For each experiment, we record the number of true positives (TP), false positives (FP), true negatives (TN), and false negatives (FN). To compare anomaly detection performance for each anomaly type, we use precision (P), recall (R), and F1 score as evaluation metrics: $\textrm{P} = \frac{\text{TP}}{\text{TP}+\text{FP}}$, $\textrm{R} = \frac{\text{TP}}{\text{TP}+\text{FN}}$ and $\text{F}_1 = \frac{2*\text{precision}*\text{recall}}{\text{precision}+\text{recall}}$.
The parameters for all models are tuned to achieve the best $F_1$ score performance. 

\subsection{Implementation Details}
Models are trained using the Adam optimizer with a learning rate of $1*10^{-3}$ for all datasets.
All training is done within 120 epochs with a batch size of 128. Other hyperparameters are shown in Table \ref{tab:hyper}. At the testing stage, each testing trajectory was randomly selected from a large data pool and the process was repeated 5 times to derive the mean values. All experiments were performed on a high-performance computing system consisting of Nvidia A100 GPUs.
\subsection{Evaluation on Baselines}
We compare our proposed method IHiD with a set of baseline models under three types of trajectory anomalies: detour, route-switching and small detour. As shown in Table \ref{tab:main results}, the results demonstrate that our method consistently achieves the best performance in terms of the $F_1$ score across all settings. 

In the Chengdu dataset, our approach brings improvements in $F_1$ score up to $15.7\%$ , $15.9\%$ and $30.2\%$
for big detour anomalies, small detour anomalies, and route-switching anomalies, respectively. GM-VASE achieves the second best performance for detecting big detour anomalies in the Chengdu dataset, while it has relatively poor performance when it comes to more subtle anomalies such as small detour anomalies and route-switching anomalies. Deep learning based methods (LSTM-AE, OIL-AD and DeCoRTAD) have good performance for detour anomalies, but less robust to detect route-switching anomalies.
In the AIS dataset, our method still leads with an $F_1$ score of $91.1\%$ for big detour anomalies, $89.1\%$ for small detour anomalies and $83.1\%$ for route-switching anomalies. In particular, traditional unsupervised baselines, such as OCSVM, LOF, and IF struggle in most cases due to their limited capacity to model complex spatiotemporal dependencies.

Among all the settings, route-switching anomalies in the AIS dataset are the most challenging to detect.
This challenge can be attributed to the following reasons:
(1) \textbf{Dataset characteristics}: A key distinction between the road traffic dataset (Chengdu) and the vessel trajectory dataset (AIS) lies in their spatial constraints. The Chengdu taxi trajectories are more constrained. In contrast, the AIS dataset represents movement in free space, where normal trajectories can exhibit substantial variability — even among those with similar origin-destination pairs.
(2) \textbf{Methodological limitations in the AIS setting}:
Our method relies on graph-based trajectory segmentation, where each node represents a region of space. In the AIS dataset, due to the unbounded nature of maritime space, regions of subgoal nodes must cover a broader area that might be of irregular shapes as well, and this leads to greater spatial ambiguity. As a result, subgoal identification might not align with the trajectories in reality, and reduce the accuracy of the high-level IQL model. 
In comparison, the Chengdu dataset uses a very small radius to define regions for the nodes, and therefore has more precise node localisation.

\subsection{Ablation Studies}

As shown in Table \ref{tab:ablation}, we further investigate the effect of the hierarchical structure in our model across different types of anomalies in the Chengdu and AIS datasets. We test the anomaly detection performance purely based on the high-level model and the low-level model separately and keep the rest of the setting the same. The results show that our approach, by using a hierarchical structure, outperforms each individual model and brings a $18.4\%$ increase in terms of the average $F_1$ score.
The high-level IQL model achieves strong performance in detecting big detours and route-switching anomalies. However, it fails to detect small detours because the high-level model only evaluates the choice of subgoals and is not able to detect deviations that occur between two subgoal nodes.
In contrast, the low-level diffusion model alone is effective in detecting small deviations (e.g. $96.0\%$ in Chengdu) but struggles with more strategic anomalies such as route switching. 
The hierarchical model,  which integrates both components, has the most balanced and robust performance across all anomaly types, with the highest average $F_1$ score of $90.7\%$.

These results demonstrate the complementary strengths of the two components: The IQL module captures high-level intent deviations, while the diffusion model helps to detect fine-grained trajectory variations. Together, they enable more comprehensive and accurate trajectory anomaly detection.

\subsection{Parameter Sensitivity}
\begin{figure}[t]
    \centering
    \subfloat{%
       \includegraphics[width=0.48\textwidth]{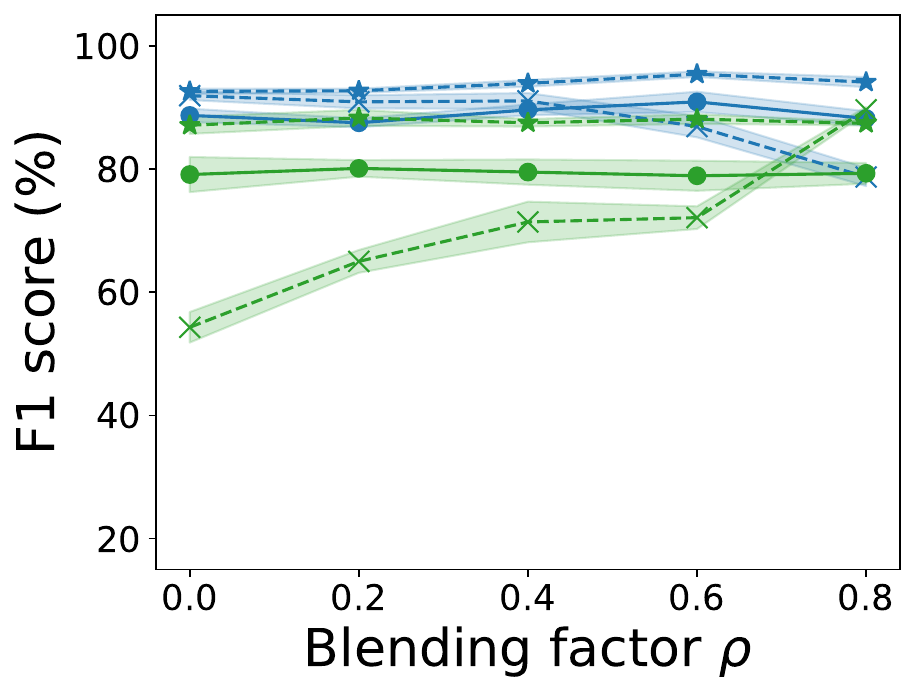} }
    \subfloat{%
       \includegraphics[width=0.48\textwidth]{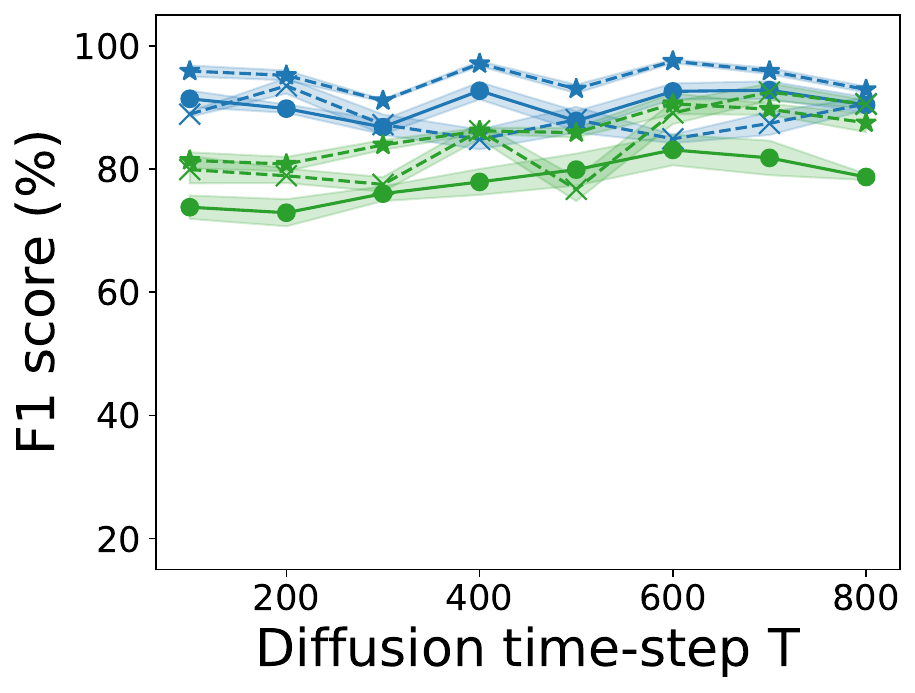}} \\
       \subfloat{%
     \includegraphics[width=0.48\textwidth]{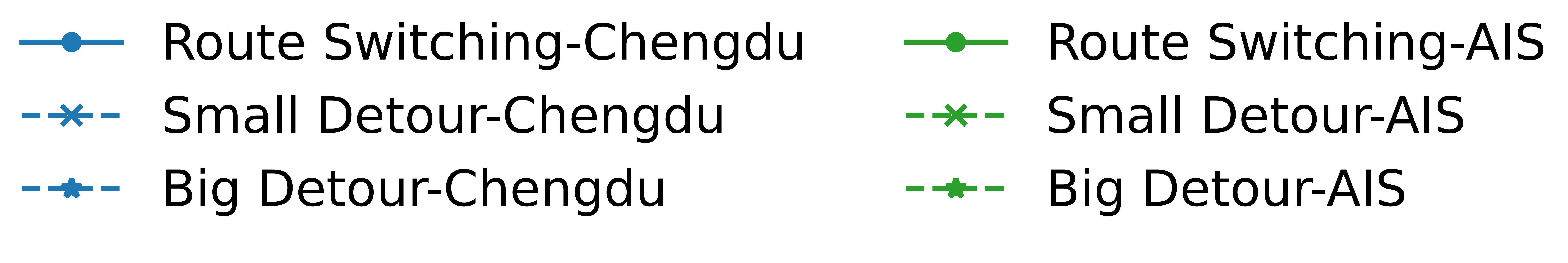}}
    \caption{Parameter sensitivity for blending factor $\rho$ (left) and Diffusion time-step $T$ (right).}
    \label{fig:parameter}
\end{figure}
We report the model's performance under varying values of two key parameters: the blending factor $\rho$ and the diffusion time-step $T$, as shown in Figure \ref{fig:parameter}. The blending factor $\rho$ primarily affects the detection of small detour anomalies. For instance, increasing $\rho$ improves the $F_1$ scores for small detour anomalies in the AIS dataset but leads to a performance drop in the Chengdu dataset.
A higher diffusion time-step $T$ (e.g., from 600 to 800) yields better performance on the AIS dataset, whereas its impact on the Chengdu dataset is less consistent. These differences can be attributed to the nature of the datasets. In the AIS dataset, which captures movement in free space, normal trajectories exhibit greater variability. This requires the diffusion model to rely more on contextual subgoal information and potentially to use higher time-step $T$ for more accurate reconstruction. In contrast, the Chengdu dataset involves road traffic trajectories that are more constrained by the underlying road topology.
\section{Conclusion}
In this paper, we address the limitations of existing long-term trajectory anomaly detection methods by introducing the Intention-aware Hierarchical Diffusion model (IHiD), a novel unsupervised framework that integrates high-level intention reasoning with low-level trajectory generation. By combining inverse Q learning for subgoal evaluation and diffusion modelling for subtrajectory reconstruction, IHiD captures both the agent’s goal-oriented behaviour and the spatial-temporal details of the trajectory movement. Furthermore, this hierarchical design allows IHiD to account for the diversity of long-term trajectories and the feasibility of implementation in real-world anomaly detection scenarios.

For future work, we will consider improving the model to generate more realistic trajectories, such as incorporating constraints of physics and road networks into our objective function.

\bibliographystyle{unsrt}  
\bibliography{references}  

\begin{thebibliography}{10}

\bibitem{saleh2018long}
Khaled Saleh, Mohammed Hossny, and Saeid Nahavandi.
\newblock Long-term recurrent predictive model for intent prediction of pedestrians via inverse reinforcement learning.
\newblock In {\em 2018 Digital Image Computing: Techniques and Applications (DICTA)}, pages 1--8. IEEE, 2018.

\bibitem{zhang2024long}
Zhiwen Zhang, Zipei Fan, Zewu Lv, Xuan Song, and Ryosuke Shibasaki.
\newblock Long-term vessel trajectory imputation with physics-guided diffusion probabilistic model.
\newblock In {\em Proceedings of the 30th ACM SIGKDD Conference on Knowledge Discovery and Data Mining}, pages 4398--4407, 2024.

\bibitem{yao2022trajgat}
Di~Yao, Haonan Hu, Lun Du, Gao Cong, Shi Han, and Jingping Bi.
\newblock Trajgat: A graph-based long-term dependency modeling approach for trajectory similarity computation.
\newblock In {\em Proceedings of the 28th ACM SIGKDD conference on knowledge discovery and data mining}, pages 2275--2285, 2022.

\bibitem{liang2023hierarchical}
Hebin Liang, Zibin Dong, Yi~Ma, Xiaotian Hao, Yan Zheng, and Jianye Hao.
\newblock A hierarchical imitation learning-based decision framework for autonomous driving.
\newblock In {\em Proceedings of the 32nd ACM International Conference on Information and Knowledge Management}, pages 4695--4701, 2023.

\bibitem{xu2023bits}
Danfei Xu, Yuxiao Chen, Boris Ivanovic, and Marco Pavone.
\newblock Bits: Bi-level imitation for traffic simulation.
\newblock In {\em 2023 IEEE International Conference on Robotics and Automation (ICRA)}, pages 2929--2936. IEEE, 2023.

\bibitem{seo2024idil}
Sangwon Seo and Vaibhav Unhelkar.
\newblock Idil: Imitation learning of intent-driven expert behavior.
\newblock In {\em Proceedings of the 23rd International Conference on Autonomous Agents and Multiagent Systems}, pages 1673--1682, 2024.

\bibitem{zhang2011ibat}
Daqing Zhang, Nan Li, Zhi-Hua Zhou, Chao Chen, Lin Sun, and Shijian Li.
\newblock ibat: detecting anomalous taxi trajectories from gps traces.
\newblock In {\em Proceedings of the 13th international conference on Ubiquitous computing}, pages 99--108, 2011.

\bibitem{chen2013iboat}
Chao Chen, Daqing Zhang, Pablo~Samuel Castro, Nan Li, Lin Sun, Shijian Li, and Zonghui Wang.
\newblock iboat: Isolation-based online anomalous trajectory detection.
\newblock {\em IEEE Transactions on Intelligent Transportation Systems}, 14(2):806--818, 2013.

\bibitem{liu2013density}
Zhipeng Liu, Dechang Pi, and Jinfeng Jiang.
\newblock Density-based trajectory outlier detection algorithm.
\newblock {\em Journal of Systems Engineering and Electronics}, 24(2):335--340, 2013.

\bibitem{han2022deeptea}
Xiaolin Han, Reynold Cheng, Chenhao Ma, and Tobias Grubenmann.
\newblock Deeptea: Effective and efficient online time-dependent trajectory outlier detection.
\newblock {\em Proceedings of the VLDB Endowment}, 15(7):1493--1505, 2022.

\bibitem{liu2020online}
Yiding Liu, Kaiqi Zhao, Gao Cong, and Zhifeng Bao.
\newblock Online anomalous trajectory detection with deep generative sequence modeling.
\newblock In {\em 2020 IEEE 36th International Conference on Data Engineering (ICDE)}, pages 949--960. IEEE, 2020.

\bibitem{zeng2021hierarchical}
Xianlin Zeng, Yalong Jiang, Wenrui Ding, Hongguang Li, Yafeng Hao, and Zifeng Qiu.
\newblock A hierarchical spatio-temporal graph convolutional neural network for anomaly detection in videos.
\newblock {\em IEEE Transactions on Circuits and Systems for Video Technology}, 33(1):200--212, 2021.

\bibitem{kong2024unbiased}
Xiangjie Kong, Yuwei He, Guojiang Shen, Jiaxin Du, Zhi Liu, and Ivan Lee.
\newblock Unbiased anomalous trajectory detection with hierarchical sequence modeling.
\newblock {\em IEEE Transactions on Consumer Electronics}, 2024.

\bibitem{wang2025oil}
Chen Wang, Sarah Erfani, Tansu Alpcan, and Christopher Leckie.
\newblock Oil-ad: An anomaly detection framework for decision-making sequences.
\newblock {\em Pattern Recognition}, page 111656, 2025.

\bibitem{jain2024go}
Abhinav Jain and Vaibhav Unhelkar.
\newblock Go-dice: Goal-conditioned option-aware offline imitation learning via stationary distribution correction estimation.
\newblock In {\em Proceedings of the AAAI conference on artificial intelligence}, volume~38, pages 12763--12772, 2024.

\bibitem{garg2021iq}
Divyansh Garg, Shuvam Chakraborty, Chris Cundy, Jiaming Song, and Stefano Ermon.
\newblock Iq-learn: Inverse soft-q learning for imitation.
\newblock {\em Advances in Neural Information Processing Systems}, 34:4028--4039, 2021.

\bibitem{ho2020denoising}
Jonathan Ho, Ajay Jain, and Pieter Abbeel.
\newblock Denoising diffusion probabilistic models.
\newblock {\em Advances in Neural Information Processing Systems}, 33:6840--6851, 2020.

\bibitem{rombach2022high}
Robin Rombach, Andreas Blattmann, Dominik Lorenz, Patrick Esser, and Bj{\"o}rn Ommer.
\newblock High-resolution image synthesis with latent diffusion models.
\newblock In {\em Proceedings of the IEEE/CVF Conference on Computer Vision and Pattern Recognition}, pages 10684--10695, 2022.

\bibitem{li2024difftad}
Chaoneng Li, Guanwen Feng, Yunan Li, Ruyi Liu, Qiguang Miao, and Liang Chang.
\newblock Difftad: Denoising diffusion probabilistic models for vehicle trajectory anomaly detection.
\newblock {\em Knowledge-Based Systems}, 286:111387, 2024.

\bibitem{shen2020timeseries}
Lifeng Shen, Zhuocong Li, and James Kwok.
\newblock Timeseries anomaly detection using temporal hierarchical one-class network.
\newblock {\em Advances in Neural Information Processing Systems}, 33:13016--13026, 2020.

\bibitem{li2021multivariate}
Zhihan Li, Youjian Zhao, Jiaqi Han, Ya~Su, Rui Jiao, Xidao Wen, and Dan Pei.
\newblock Multivariate time series anomaly detection and interpretation using hierarchical inter-metric and temporal embedding.
\newblock In {\em Proceedings of the 27th ACM SIGKDD conference on knowledge discovery \& data mining}, pages 3220--3230, 2021.

\bibitem{kong2020huad}
Xiangjie Kong, Haoran Gao, Osama Alfarraj, Qichao Ni, Chaofan Zheng, and Guojiang Shen.
\newblock Huad: Hierarchical urban anomaly detection based on spatio-temporal data.
\newblock {\em IEEE Access}, 8:26573--26582, 2020.

\bibitem{pan2020xgail}
Menghai Pan, Weixiao Huang, Yanhua Li, Xun Zhou, and Jun Luo.
\newblock xgail: Explainable generative adversarial imitation learning for explainable human decision analysis.
\newblock In {\em Proceedings of the 26th ACM SIGKDD International Conference on Knowledge Discovery \& Data Mining}, pages 1334--1343, 2020.

\bibitem{feng2023ilroute}
Tao Feng, Huan Yan, Huandong Wang, Wenzhen Huang, Yuyang Han, Hongsen Liao, Jinghua Hao, and Yong Li.
\newblock Ilroute: A graph-based imitation learning method to unveil riders' routing strategies in food delivery service.
\newblock In {\em Proceedings of the 29th ACM SIGKDD Conference on Knowledge Discovery and Data Mining}, pages 4024--4034, 2023.

\bibitem{oh2019sequential}
Min-hwan Oh and Garud Iyengar.
\newblock Sequential anomaly detection using inverse reinforcement learning.
\newblock In {\em Proceedings of the 25th ACM SIGKDD International Conference on Knowledge Discovery \& data mining}, pages 1480--1490, 2019.

\bibitem{jing2021adversarial}
Mingxuan Jing, Wenbing Huang, Fuchun Sun, Xiaojian Ma, Tao Kong, Chuang Gan, and Lei Li.
\newblock Adversarial option-aware hierarchical imitation learning.
\newblock In {\em International Conference on Machine Learning}, pages 5097--5106. PMLR, 2021.

\bibitem{wyatt2022anoddpm}
Julian Wyatt, Adam Leach, Sebastian~M Schmon, and Chris~G Willcocks.
\newblock Anoddpm: Anomaly detection with denoising diffusion probabilistic models using simplex noise.
\newblock In {\em Proceedings of the IEEE/CVF Conference on Computer Vision and Pattern Recognition}, pages 650--656, 2022.

\bibitem{pintilie2023time}
Ioana Pintilie, Andrei Manolache, and Florin Brad.
\newblock Time series anomaly detection using diffusion-based models.
\newblock In {\em Proceedings of the 2023 IEEE International Conference on Data Mining Workshops (ICDMW)}, pages 570--578. IEEE, 2023.

\bibitem{liu2023unsupervised}
Zhenzhen Liu, Jin~Peng Zhou, Yufan Wang, and Kilian~Q Weinberger.
\newblock Unsupervised out-of-distribution detection with diffusion inpainting.
\newblock In {\em Proceedings of the International Conference on Machine Learning}, pages 22528--22538. PMLR, 2023.

\bibitem{flaborea2023multimodal}
Alessandro Flaborea, Luca Collorone, Guido Maria~D'Amely di~Melendugno, Stefano D'Arrigo, Bardh Prenkaj, and Fabio Galasso.
\newblock Multimodal motion conditioned diffusion model for skeleton-based video anomaly detection.
\newblock In {\em Proceedings of the IEEE/CVF International Conference on Computer Vision}, pages 10318--10329, 2023.

\bibitem{livernoche2023diffusion}
Victor Livernoche, Vineet Jain, Yashar Hezaveh, and Siamak Ravanbakhsh.
\newblock On diffusion modeling for anomaly detection.
\newblock In {\em Proceedings of the Twelfth International Conference on Learning Representations}, 2023.

\bibitem{xiao2023imputation}
Chunjing Xiao, Zehua Gou, Wenxin Tai, Kunpeng Zhang, and Fan Zhou.
\newblock Imputation-based time-series anomaly detection with conditional weight-incremental diffusion models.
\newblock In {\em Proceedings of the 29th ACM SIGKDD Conference on Knowledge Discovery and Data Mining}, pages 2742--2751, 2023.

\bibitem{wang2024decortad}
Chen Wang, Sarah Erfani, Tansu Alpcan, and Christopher Leckie.
\newblock Decortad: Diffusion based conditional representation learning for online trajectory anomaly detection.
\newblock In {\em ECAI 2024}, pages 2757--2764. IOS Press, 2024.

\bibitem{ziebart2008maximum}
Brian~D Ziebart, Andrew~L Maas, J~Andrew Bagnell, and Anind~K Dey.
\newblock Maximum entropy inverse reinforcement learning.
\newblock In {\em Proceedings of the AAAI}, pages 1433--1438, 2008.

\bibitem{fu2017learning}
Justin Fu, Katie Luo, and Sergey Levine.
\newblock Learning robust rewards with adversarial inverse reinforcement learning.
\newblock {\em arXiv preprint arXiv:1710.11248}, 2017.

\bibitem{ronneberger2015u}
Olaf Ronneberger, Philipp Fischer, and Thomas Brox.
\newblock U-net: Convolutional networks for biomedical image segmentation.
\newblock In {\em Proceedings of the 18th International Conference on Medical Image Computing and Computer-Assisted Intervention--MICCAI 2015}, pages 234--241, 2015.

\bibitem{chen2023executing}
Xin Chen, Biao Jiang, Wen Liu, Zilong Huang, Bin Fu, Tao Chen, and Gang Yu.
\newblock Executing your commands via motion diffusion in latent space.
\newblock In {\em Proceedings of the IEEE/CVF Conference on Computer Vision and Pattern Recognition}, pages 18000--18010, 2023.

\bibitem{kipf2019compile}
Thomas Kipf, Yujia Li, Hanjun Dai, Vinicius Zambaldi, Alvaro Sanchez-Gonzalez, Edward Grefenstette, Pushmeet Kohli, and Peter Battaglia.
\newblock Compile: Compositional imitation learning and execution.
\newblock In {\em International Conference on Machine Learning}, pages 3418--3428. PMLR, 2019.

\bibitem{didi}
{DiDi GAIA Open Dataset}.
\newblock \url{https://outreach.didichuxing.com/en/}, 2016.

\bibitem{AIS}
{Australian Maritime Safety Authority Digital Data}.
\newblock \url{https://www.operations.amsa.gov.au/Spatial/DataServices/DigitalData}, 2020.

\bibitem{malhotra2016lstm}
Pankaj Malhotra, Anusha Ramakrishnan, Gaurangi Anand, Lovekesh Vig, Puneet Agarwal, and Gautam Shroff.
\newblock Lstm-based encoder-decoder for multi-sensor anomaly detection.
\newblock {\em arXiv preprint arXiv:1607.00148}, 2016.

\end{thebibliography}


\end{document}